# Tensegrity crutches with compliance from a pre-stressed self-tensile module improve ground reaction force profiles, speed, effort, comfort, and perceived stability

Jingxian Gu[1], Joanna Spyra[2], Andrew Walski[1], Lyla Elsaesser[1], Samuel Bierner[3], Dobromir Dotov[1*]

[1] Department of Biomechanics, University of Nebraska Omaha, Omaha, NE, USA

[2] Independent Researcher, Omaha, Nebraska

[3] Department of Physical Medicine and Rehabilitation, University of Nebraska Medical Center, Omaha, NE, USA

Jingxian Gu: ORCiD 0009-0007-3822-9926
Joanna Spyra: ORCiD 0000-0002-3585-3914
Samuel Bierner: ORCiD 0000-0001-9378-7209
Dobromir Dotov: ORCiD 0000-0002-5543-360X
* ddotov@unomaha.edu

**Purpose**: Six million people use crutches as mobile aids in the US. Rigid designs with no axial mobility limit sensory feedback and lead to secondary injury on the upper joints. Spring-loaded designs offer compliance but may compromise stability. We designed a biologically inspired tensegrity crutch with a compliant module aiming to achieve favorable mechanical properties. The terminal module was a pre-stressed self-tensile two-cell tensegrity structure. We compared the tensegrity crutch to commercial rigid and spring-loaded crutches in mechanical tests using axial loading, in overground straight and turning walking, and in participant experience.

**Methods**: In human trials, healthy young adults (N=18) with no recent lower-body injury performed straight walking and turning trials at a comfortable self-selected pace. A knee blocker simulated unilateral injury of the dominant leg. After using each type of crutch, participants reported their perceived levels of effort, comfort, pain, stability, and usability.

**Results**: Compared to the rigid design, both spring-loaded and tensegrity conditions reduced peak loading rates. The tensegrity design improved effort, comfort, pain, and usability. Spring-loaded crutches reduced perceived stability and walking speed.

**Conclusion**: The biologically inspired tensegrity crutches were an overall improvement to existing designs. Simulations and mechanical testing suggest that nonlinear stiffness, ground-following, and force feedback are among the beneficial mechanical properties that underlie this improvement.

## Introduction

Human bodies and assistive devices aimed at supporting load-bearing functions of the human body face the fundamental challenge of satisfying the conflicting requirements of mobility, stability, comfort, and sensory feedback. Mobility assistive devices (MAD's) such as canes, crutches, and walkers, are commonly used to address one's essential needs for mobility, independent living, and ability to perform daily activities. Crutches are among the primary aids for enabling ambulation in persons with a range of disabling conditions, such as amputation, spinal cord injury, or disability due to diabetes (Berardi et al., 2021; WHO & Fund (UNICEF), 2022). In the US, six million people use crutches (Rasouli & Reed, 2020a), and 24% of the population aged 65 and older reported using a mobility device within a month of responding, with one-third of these individuals using multiple assistive devices (Gell et al., 2015).

There is consistent evidence for the increased risk of secondary injury due to chronic use of crutches. Once crutches become the primary device for ambulation, the load-bearing upper-body joints become exposed to regular impacts for which they are not adapted. In contrast to the way human bodies naturally support movement, existing mobility aids for walking tend to be rigid or with very limited axial mobility. They transfer ground reaction forces directly to the upper body which leads to chronic pain, secondary injury, and neuropathy. Reported injuries associated with crutch use include brachial plexus compressive neuropathy from axillary crutches (Raikin & Froimson, 1997), ulnar nerve compression neuropathy at Guton's canal caused by crutch walking (Ginanneschi et al., 2009), peripheral neuropathies in nonparetic upper extremities of stroke patients (Dozono et al., 2015), and shoulder pain (Jain et al., 2010).

Rigidity also limits sensory feedback from ground reaction forces and contributes to impaired proprioception and increased risk of falling (Bradley & Hernandez, 2011). Walking with a crutch fundamentally changes the gait pattern compared to normal walking (Rasouli & Reed, 2020b). Notable biomechanical differences include reduced walking velocity, disrupted arm swing, increased vertical fluctuation of the shoulders, greater trunk flexion, altered ranges of joint displacement, plantar foot pressure, and ground reaction forces. Crutch-assisted gait requires more energy and higher heart rate (Hinton & Cullen, 1982). At the level of functional coordination, the impact of such assistive devices remains debatable. Some studies suggest that assistive devices improve coordination and balance during walking (Graafmans et al., 2003), while others report the opposite effect (Bateni et al., 2004). Despite known negative side effects, there has been very little practical evolution in the design of load-bearing assistive devices. To offset the risks, chronic users have no choice but to reduce their mobility (Luz et al., 2017; West et al., 2015).

Recent progress in human movement science and biologically inspired robotics points to a generic approach to designing load-bearing assistive devices that satisfy the conflicting requirements of mobility, stability, comfort, and sensory feedback. The tensegrity hypothesis states that human bodies satisfy these requirements by implementing self-tensile networks of interconnected bones, tendons, and other connective tissue (Turvey & Fonseca, 2014), Figure 1a. Tensegrities are self-tensile designs, also known as closed kinematic chains, consisting of compression elements suspended inside a pre-stressed network of tensions elements (Snelson, 2012), see Figure 1b. The compression elements maintain the volume and pre-stress of the overall solid and, paradoxically, have a certain amount of free movement inside of it. The tension elements distribute forces and structural equilibrium. The favorable compliant properties of tensegrities come from the fact that the network of connections absorbs impact by transferring it among all elements. By distributing mechanical transduction via continuous tension elements enclosing discontinuous compression elements, such pre-stressed networks achieve high strength-to-weight ratio, form-finding stability, elastic absorption of unpredictable perturbations (Bongard, 2013). Theoretically, their global instantaneous mechano-transduction favors kinesthetic and haptic perception of the environment (Turvey & Fonseca, 2014)

The evidence is mounting that biology exploits tensegrities at the level of cellular and musculo-skeletal organization (Ingber, 2003; S. Levin et al., 2017). In terms of engineering, tensegrity principles have been successfully implemented at the architectural scale of buildings and bridges (Ingber, 2003; S. Levin et al., 2017; S. M. Levin, 2002; Micheletti & Podio-Guidugli, 2022). In biologically inspired robotics, the relevance of such structures has been demonstrated through numerous proof-of-principle examples of robots and passive and active limbs (Bongard, 2013; Caluwaerts et al., 2014; Chen et al., 2017; Jung et al., n.d.; Li et al., 2020; Marjaninejad et al., 2020; Melnyk & Pitti, 2018; Mirletz et al., 2014). The designs demonstrate favorable properties such as adaptive compliance, high strength-to-weight ratio, self-balancing, and morphological computation (Bongard, 2013; Caluwaerts et al., 2013; Jung et al., 2018; Melnyk & Pitti, 2018; Mirletz et al., 2014; Paul et al., 2006; Scarr, 2012). They recover their preferred shape after perturbation (Silva et al., 2010) and the dynamics and preferred shape can be controlled at a low energy cost by adjusting tension elements (Chen et al., 2017). Altogether, this promises exciting opportunities to advance the design of real-world mobility aids for human use (Lee et al., 2021; Sun et al., 2020; Sun & Wang, 2022; Woods & Vikas, 2023).

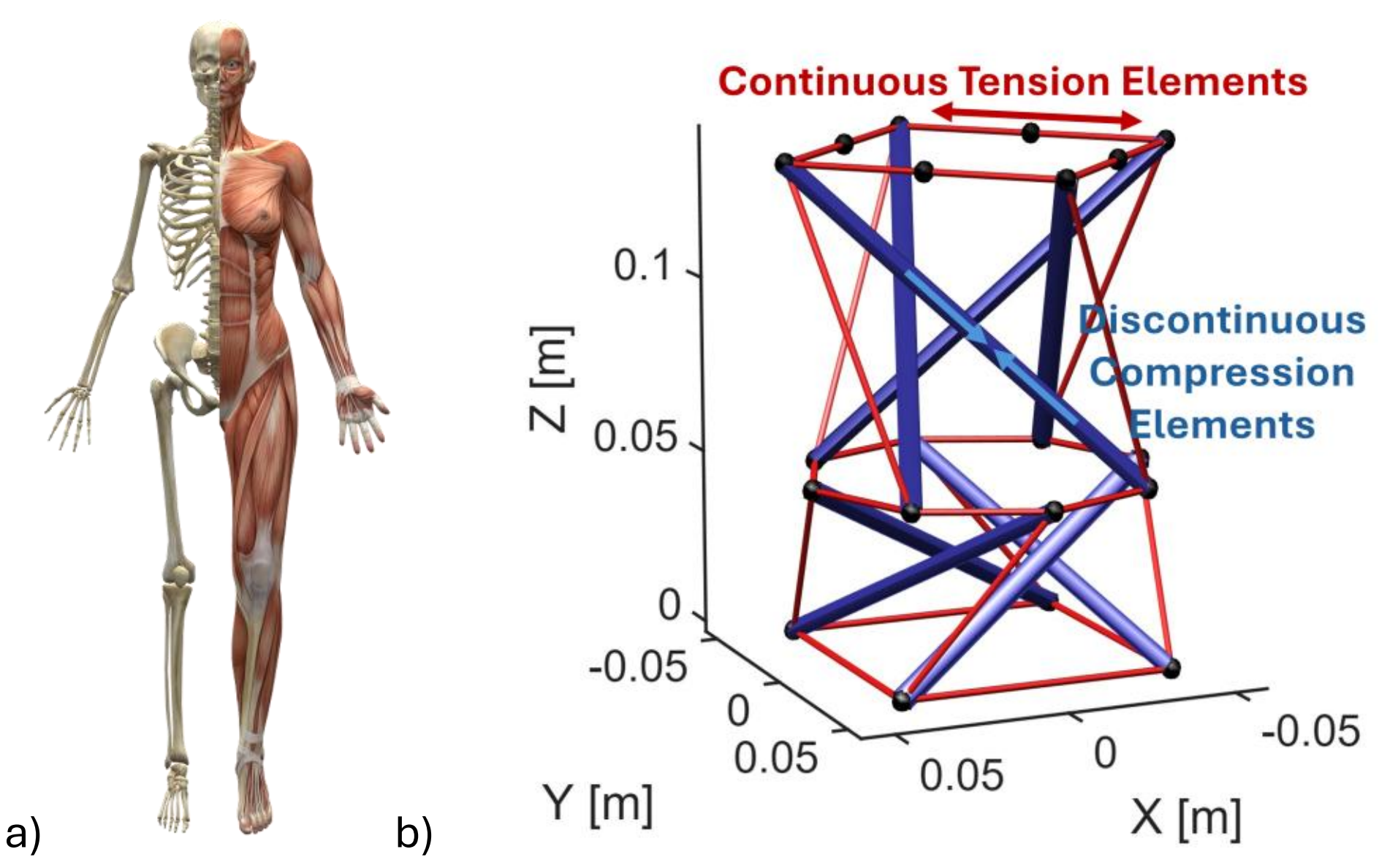

*Figure 1*. a) The human musculoskeletal system, including its tendons, can be conceptualized as a combination of compression and tension elements. b) Design of the columnar structure with multi-dimensional compliance employed in the terminal module of our compliant crutch.

The goal of this study was to implement principles of pre-stress, self-tensile networks, and multi-dimensional compliance in load-bearing mobility assistive devices, namely crutches, and to test how these principles improved the user's comfort, mobility, and stability by comparing against standard crutches. We developed a novel design and compared it against two commercially available types, a standard rigid crutch and an axially spring-loaded design with a pivoting tip. Here, we report an overground walking study on straight and turning paths and we focus the analysis on ground reaction forces, speed of walking, and user reported evaluations of relevant parameters such as effort, pain, and perceived stability.

## Materials and Methods

### Tensegrity crutch

A load-bearing mobility assistive device intended to support free dynamic interactions with the ground of an adult human poses very high demands in terms of robustness and reliability. The final design consisted of a columnar tensegrity module added at the bottom of a standard rigid crutch as a terminal ground-contact stage. The terminal stage consisted of a two-cell, four-strut columnar tensegrity made from steel cables, aluminum threaded rods, nuts, and 3D-printed plastic joint bits, see Figure 1b and 2a. Concentrating the compliance at the bottom of the structure makes it poised to conform to ground irregularities, see Figure 2a and 2c. This is consistent with observations that in humans, the ankle increases compliance whereas muscles of the more proximal joints become more activated and stiffen the joints when walking on slippery terrain (Whitmore et al., 2016). The development process was an iterative process during which the structure, materials, shape, attachment, and size of the tensegrity module were adapted after pilot tests and computational modeling. We relied on TsgFEM, a toolbox dedicated for finite element modeling of tensegrities (Ma et al., 2022). Furthermore, we consulted with stakeholders who suggested that it is important to increase device stability by way of multi-point ground contact, but the existing multi-point tips create other complications due to their rigidity.

Achieving high pre-stress is an essential property of tensegrity structures but also a major challenge to the assembly and reliability of modules that are practical for use by humans in real-world conditions. The struts (compression elements) in our design were threaded rods with nuts and movable plastic joint bits. The cables (tension elements) were hooked on the joint bits using high-strength loops. Turning the nuts had the effect of pushing the joint bits away from each other, thus expanding the effective length of the struts and applying pre-stress to the network of cables (tension elements) (Małyszko & Rutkiewicz, 2020). The four struts of the lower cell instantiated separate points of contact with the ground that were individually mobile.

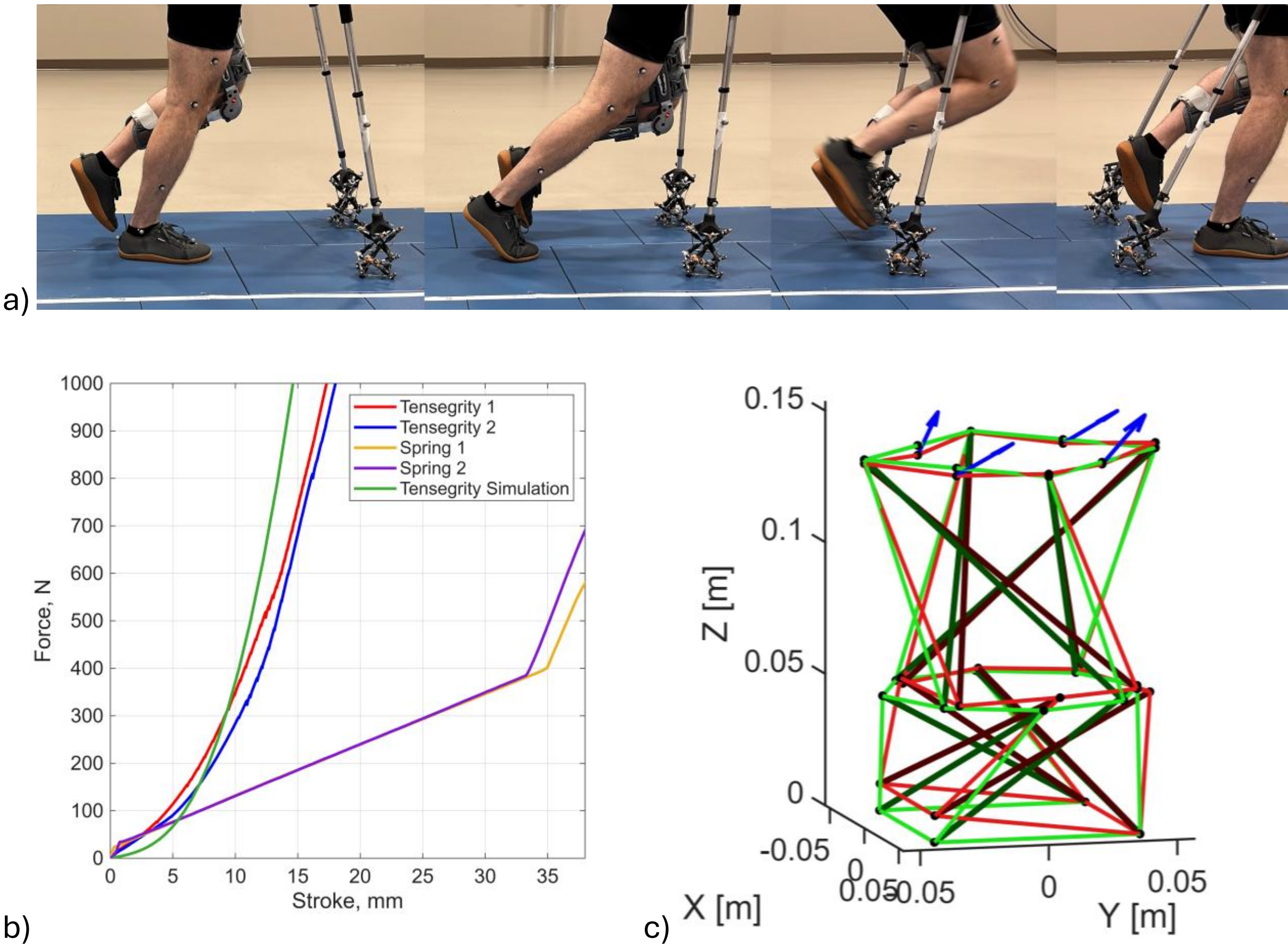

*Figure 2*. a) Compliance of the terminal modules during ambulation with tensegrity crutches. b) Stiffness profiles of two tensegrity crutches, a simulation of the same tensegrity structure, and two commercially available spring-loaded crutches. c) Simulated conformation to ground imperfections and structured force feedback. (green) unloaded state on flat ground. (red) loaded state with -500 N vertical force at the top nodes and an uneven terrain elevating two of the bottom nodes.

We conducted mechanical testing of individual components and the entire tensegrity module using axial loading on a universal testing machine (Shimadzu, Kyoto, Japan). The failure load of individual cables with loops was between 1500 and 2000 N. The failure load of the entire two-cell module was between 2000 and 4500 N and was mainly limited by the cable loops and materials used for the 3D-printed joint bits.

An important property of the tensegrity module is its nonlinear stiffness profile. At the design stage, simulations showed that the tensegrity module should exhibit relatively low stiffness with small loads and quickly increasing stiffness with larger loads. Indeed, this was confirmed by tests in the vertical axis performed with the universal testing machine. As Figure 2b shows, the force to displacement relation was a nonlinear one. At small loads, the stiffness started at 16.3 kN/m but accelerated with increasing force and it was 121.3 kN/m at around 1000 N. In contrast, the axial stiffness of the commercial spring-loaded crutch was a constant 10.8 kN/m until it traveled about 3.5 cm and bottomed out, hitting the rubber tip, see Figure 2b. Nonlinear stiffness of load-bearing devices is a desirable property because it may mimic the naturally nonlinear stiffness of human lower-body segments during walking (Grabowski et al., 2010).

**Participants**

Eighteen young healthy adults (*N* = 18; male/female = 15/3; age range= 21-33 years; left/right dominant leg = 1/17) participated in exchange for a monetary compensation. Eligibility criteria included: (1) age between 19 and 35 years; (2) ability to walk independently without the use of an assistive device; (3) normal or corrected-to-normal vision; (4) no known neurological, lower-extremity musculoskeletal, or cardiovascular disorders; (5) no current lower-limb injuries or disabilities; (6) no use of medications known to affect motor performance; and (7) no use of crutches within the previous six months. All participants provided informed consent. The study ethics approval was approved by the University of Nebraska Medical Center Institutional Review Board (IRB#0702-24-EP).

**Procedure**

Upon arrival at the laboratory, participants were screened by a researcher to confirm eligibility. Participants provided written informed consent prior to participation. Demographic information, including gender, age, height, and body mass, was collected. Participants completed a questionnaire on prior crutch use and exercise background. The crutches were then fitted. Participants were dressed in a T-shirt and singlet. They wore their comfortable athletic shoes to minimize learning effects associated with unfamiliar footwear. Forty-five reflective markers were placed on anatomical landmarks according to a full-body marker set. Instructions on the questionnaires were provided and participants were encouraged to ask questions at any time.

Participants received instructions and demonstrations by the experimenter and then practiced walking. Once having demonstrated adequate proficiency and reported confidence with all crutch types, participants practiced walking back and forth over the ground-embedded force plates while performing straight walking trials and keeping each crutch on its row of adjacent force plates. After practice, straight walking with random-order blocked trials with each crutch were performed. Followed by turning walking blocks. Breaks for seated rest were provided between each block and on demand.

**Apparatus**

Three types of crutches were used in this study: 1) rigid type, 2) spring-loaded type with a rubber pivot tip, and 3) a custom-designed tensegrity crutch. The first two types were commercially available. All had an adjustable length. For fair comparison, we fitted the same forearm (Canadian) upper part to all three types, and added masses to equalize their weight (19.2N) as the commercial spring-loaded type was inherently heavier. A knee blocker (9.81N) was used to fix the knee at 90 degrees and simulate an injury of the dominant leg. The walking pattern with two parallel crutches and one leg allowed us to isolate ground reaction forces from the crutch on the affected side.

**Experimental Design**

To standardize loading conditions, weight bearing was assigned to the non-dominant leg and the dominant was braced at the knee. Participants walked in a two-point gait with alternating support between one leg and two crutches. The dominant leg was fixed in a flexed position with a knee brace and kept off the ground at all time. During ambulation, participants were instructed to advance the crutches approximately 5 cm lateral to the foot and 30 cm anterior to the body, then step forward with the weight-bearing lower limb while pushing down through with the handgrips. Emphasis was placed on avoiding excessive medial and lateral placement of the crutches and maintain appropriate spacing between crutches tips.

The walking task consisted of overground ambulation at the participants' self-selected comfortable speed either on a straight back and forth path or on a path with a ninety degrees turn. In straight walking trials, only the straight portion over the force plates was recorded. In turning trials, the path involved a single ninety degrees turn and participants walked back and forth on a path marked with cones on the ground. Following completion of each block of trials, participants reported their evaluation of the crutches they used in that block.

**Data Collection**

Ground reaction forces were sampled at 1000 Hz with force plates embedded in the floor (Optima High Performance Series, AMTI Inc., Watertown, MA, USA). Seven force plates (40×60 cm) were arranged in a modular array covering an area of 2.4x0.8 meters, including false plates filling in extra spaces to minimize visual cueing. This arrangement allowed us to record multiple gait cycles per pass. Kinematic data were recorded using 12 infrared cameras (Kestrel 4200TM, Motion Analysis Corp., Rohnert Park, CA, USA) at 100 Hz.

The questionnaire included standard scales and custom Likert scales: perceived effort Borg scale (0-10 from rest to extremely hard) (Borg, 1982), comfort level (-3 to 3 from very uncomfortable to very comfortable), perceived stability (-3 to 3 from very unstable to very stable), pain level (0 to 10, from no pain to worst pain possible), and system usability scale (SUS, 0-100) (Brooke, 1996). Participants also provided optional written comments. The questionnaire was completed after each block.

**Data Processing**

Crutch-strike events were identified in the vertical ground reaction force signals of each force plate (vGRF) by applying a threshold window of minimum 20 N and minimum maintenance of 10 ms. Valid crutch stance cycles were retained following the criteria of stance duration (at least have 400 ms), peak vGRF greater than one third of body weight, low to high rising vGRF within the first 10% of crutch stance phase. Force data were filtered using a fourth order Butterworth low pass filter. The stance phase of the crutch was time-stretched to a common time-frame expressed as percent stance phase. Forces were normalized to body weight.

Loading rates were calculated in the initial 200 ms after each crutch strike and the peak was the maximum loading rate in any 10 ms time-bin. Impulse was computed by integrating the force profile over stance phase. Anteroposterior impulse was separated into braking and propulsive components based on the sign of the force profile. For walking speed, the horizontal displacements of a dorsal marker in the principal axis of the walking path were differenced and averaged. The rest of the kinematics will be reported separately.

**Statistical Analysis**

Linear mixed effects models were used for statistical analysis (lme4 for R). This approach provides flexibility in handling unbalanced designs and is more robust to violations of assumptions associated with traditional statistical methods. Block and trial number were included as fixed effects to account for potential time related effects such as fatigue or learning. All model included a random intercept to account for inter-individual variability in baseline. We followed a minimal to maximal regression model procedure by iteratively adding fixed effects until we found the largest model with significant improvement in goodness-of-fit.

The maximal possible model structure for kinetic variables was specified as: $Y_{ij} = \beta_0 + \sigma_0 i + \beta_1 \text{Device}_{ij} + \beta_2 \text{Block}_{ij} + \beta_3 \text{Trial}_{ij} + \beta_4 \text{Device}_{ij}\text{Trial}_{ij} + \sigma_{ij}$, where $i$ is participant, $j$ is trial, $\sigma_0 i$ is individual baseline variability, $\sigma_{ij}$ is the residual variability. For the kinematic variable of speed, straight and turning walking were analyzed as separate variables. The maximal possible model was: $Y_{ij} = \beta_0 + \sigma_0 i + \beta_1 \text{Device}_{ij} + \beta_2 \text{Block}_{ij} + \beta_3 \text{Trial}_{ij} + \beta_4 \text{Device}_{ij}\text{Trial}_{ij} + \beta_5 \text{Device}_{ij}\text{Block}_{ij} + \sigma_{ij}$, where $i$ is participant, $j$ is trial, $\sigma_0 i$ is individual baseline variability, $\sigma_{ij}$ is the residual variability. For self-reported evaluations of the different crutch types, the maximal model structure was: $Y_{ij} = \beta_0 + \sigma_0 i + \beta_1 \text{Device}_{ij} + \beta_2 \text{Turning}_{ij} + \beta_3 \text{Device}_{ij}\text{Turning}_{ij} + \sigma_{ij}$, where $i$ is participant, $j$ is trial, $\sigma_{0,i}$ is individual baseline variability, $\sigma_{ij}$ is the residual variability.

The full statistical models for each dependent variable reported here are included in Appendix A. The tables include the sequences from minimal to maximal model, their estimated parameters and goodness-of-fit measures. In the main text, we report only the effects from the final selected model.

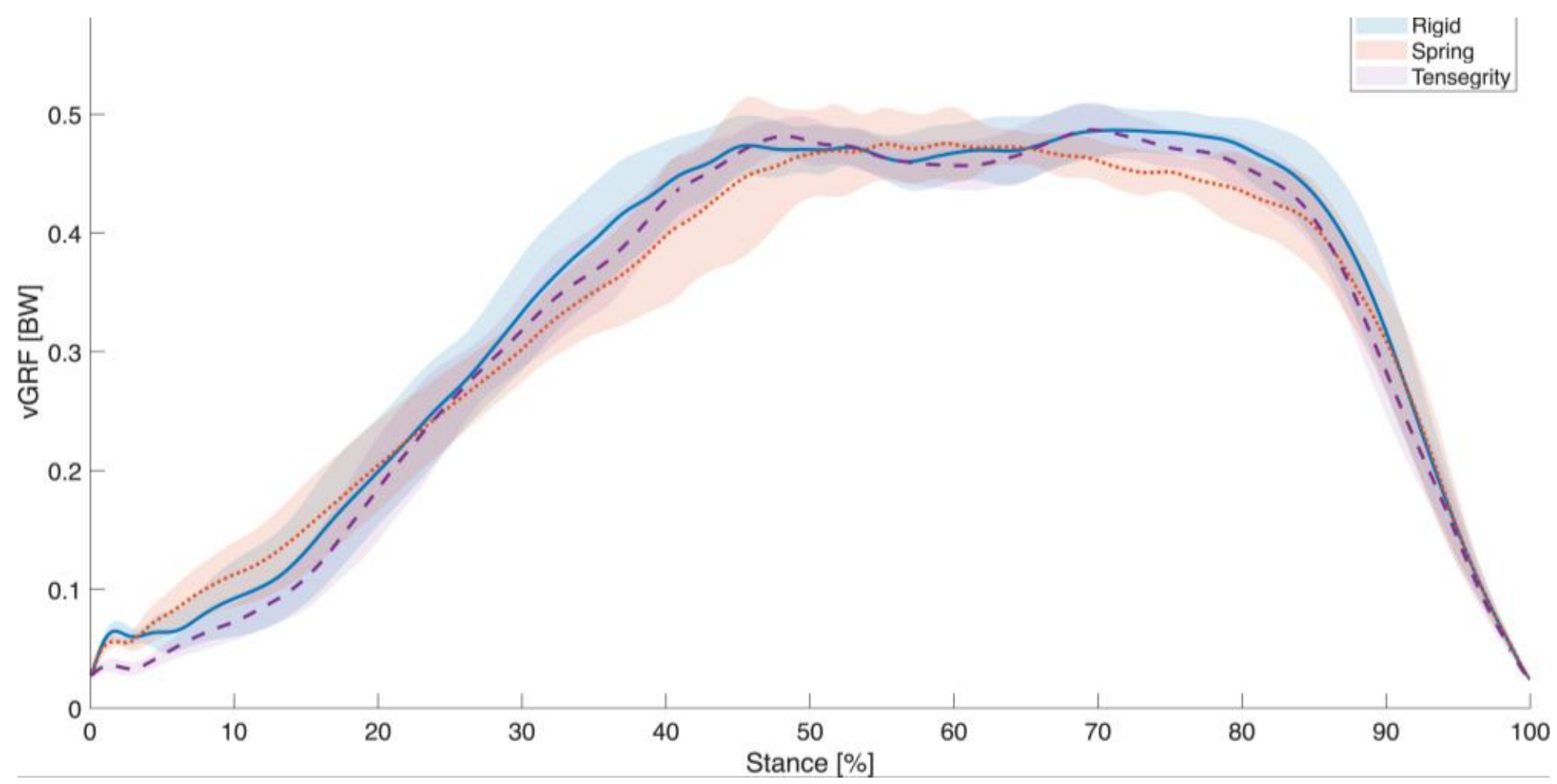

*Figure 3*. Representative vertical ground reaction force profiles from one participant are shown averaged (±*SD*) per condition after normalizing time along the stance cycle.

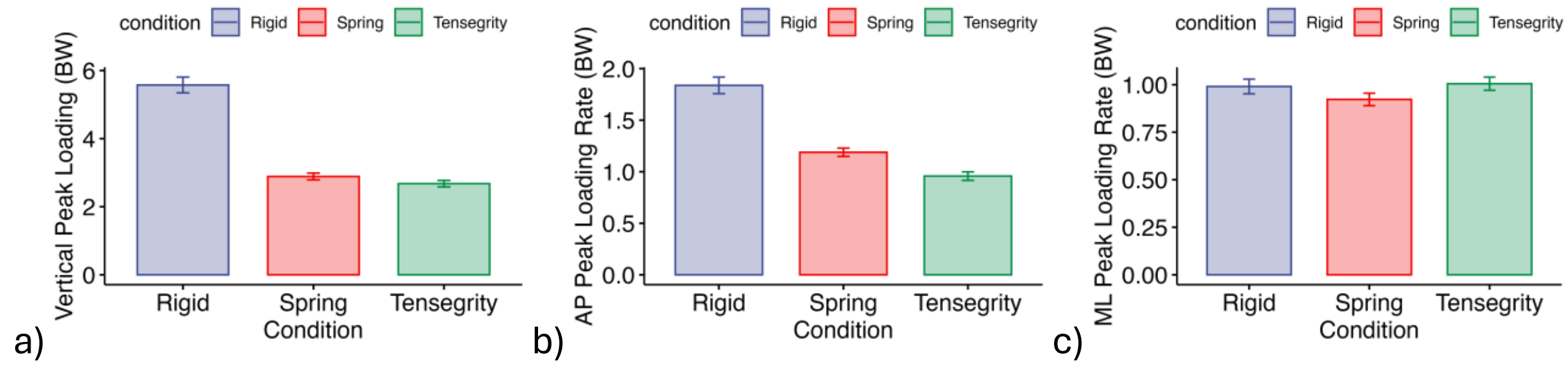

*Figure 4*. Summary statistics (*M*±*SE*) of peak loading rate immediately after ground contact. a) In the vertical axis, peak loading rate was highest with the rigid crutches, followed by the spring-loaded and tensegrity types. b) A similar pattern was observed in the anterior-posterior (front-back) axis where rigid was highest, spring-loaded was lower than rigid, and tensegrity was lower than spring-loaded. c) There were no differences in the mediolateral axis.

## Results

### Gait Kinetics

#### *Peak loading rate*

Compared to rigid crutches, both spring-loaded ($\beta = -2.7347$, SE = 0.1817, $p < 2e-16$) and tensegrity crutches ($\beta = -3.0931$, SE = 0.1854, $p < 2e-16$) significantly reduced peak loading rate in the vertical direction. There was no significant difference between spring-loaded crutches and tensegrity crutches (β = 0.358, SE = 0.190, p = 0.1432), see Figure 4a. In the anteroposterior dimension, compared to rigid crutches, both spring-loaded (β = -0.69161, SE = 0.06830, p < 2e-16) and tensegrity crutches (β = -0.96667, SE = 0.06968, p < 2e-16) significantly reduced peak loading rate. The device effect size was large ($\eta^2$ partial= 0.22). Pairwise comparisons indicated that all conditions differed significantly, with the tensegrity crutch lower than spring-loaded (p < .05), which was lower than rigid (p < .05), see Figure 4b. No significant effects of block or trial number were observed. In the mediolateral direction, there were no significant effects of device, see Figure 4c. Significant time-related effects were observed. Peak loading rate increased across blocks (β = 0.062471, SE = 0.021513, p = 0.0038, $\eta^2$ partial = 0.01), while it decreased with trial number (β = -0.011248, SE = 0.005510, p= 0.0416, $\eta^2$ partial = 5.75e-03).

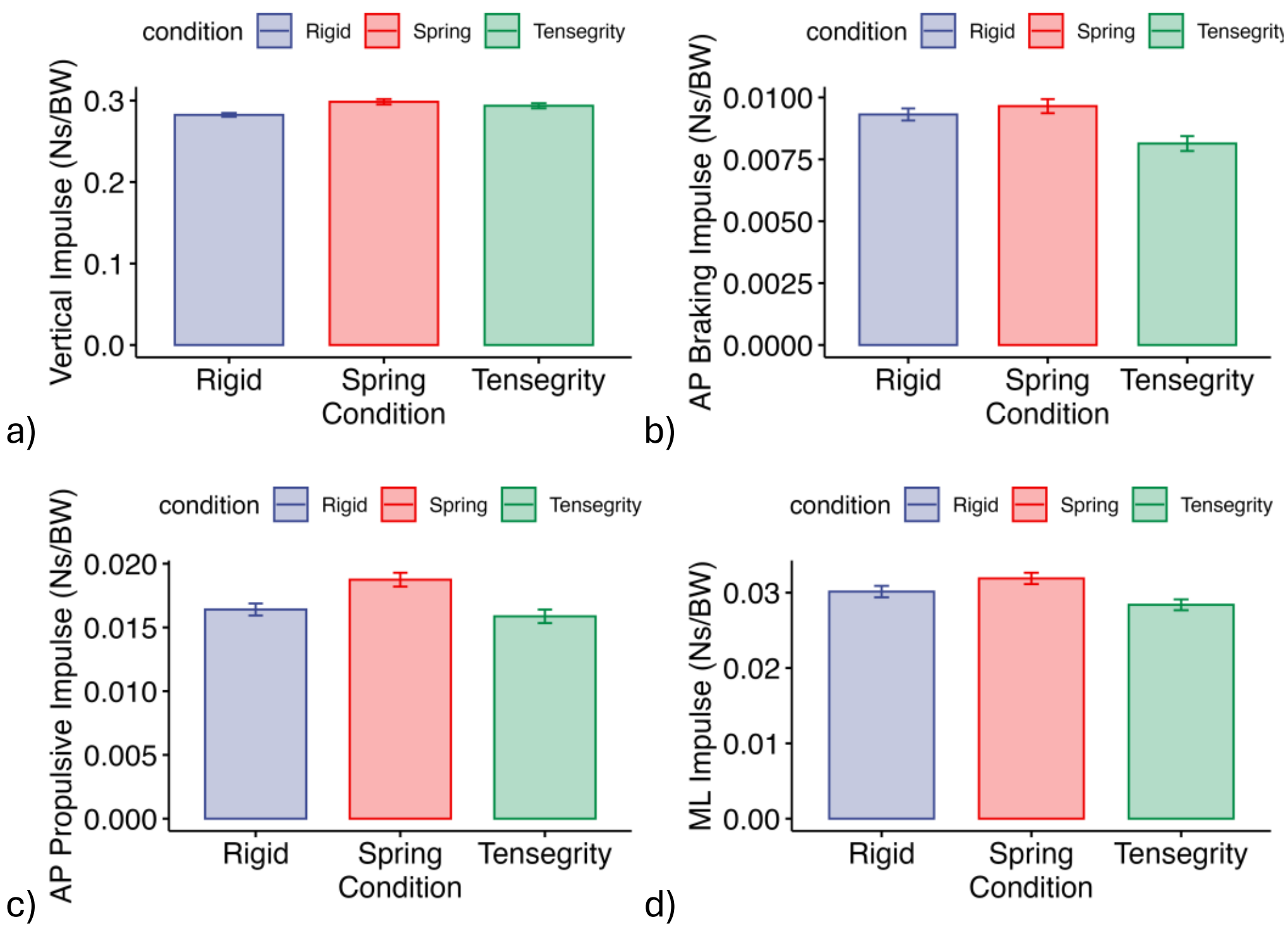

*Figure 5*. Impulse normalized by body weight (Ns/BW) summarized (*M*±*SE*) per device in the a) vertical axis, b) AP axis in the breaking phase, c) AP axis in the propulsion phase, and d) ML axis.

### *Impulse*

In the vertical axis, in comparison to the rigid crutches, impulse was significantly higher when using spring-loaded crutches (β = 0.015851, SE = 0.002956, p <.0001, $\eta^2$ = 0.04) and tensegrity crutches (β = 0.008943, SE = 0.003155, p <0.01), see Figure 5a. Pairwise comparisons

further indicated that there was no significant difference between tensegrity and spring ($\beta$ = 0.00691, SE = 0.00313, p =0.0708). Vertical impulse tended to decrease with consecutive blocks ($\beta$ = -0.005938, SE = 0.001556, p <.001, $\eta^2$ = 0.02), suggestive of ongoing adaptation or learning occurring within device.

In the AP axis in the breaking phase, impulse was significantly lower when using tensegrity crutches compared to rigid crutches ($\beta$ = -1.215e-03, SE = 2.921e-04, p = 3.59e-05, $\eta^2$ = 0.04), while there was no significant difference between the rigid and spring-loaded crutches ($\beta$ = 2.327e-04, SE = 2.801e-04, p = 0.406), see Figure 5b. In the AP axis in the propulsion phase, there was a significant difference between rigid and spring-loaded devices ($\beta$ = 2.724e-03, SE = 6.007e-04, p = 6.75e-06, $\eta^2$ = 0.03), and there was no significant difference between rigid and tensegrity devices ($\beta$ = 1.629e-04, SE = 6.410e-04, p = 0.799452), see Figure 5c. Pairwise comparisons further indicated that the propulsion impulse was higher with the spring-loaded than with the tensegrity crutches ($\beta$ = 0.002561, SE = 0.000636, p = 0.0002). Furthermore, propulsive impulse tended to increase along blocks ($\beta$ = 6.721e-04, SE = 3.165e-04, p = 0.034061, $\eta^2$ = 6.23e-03) and trial number ($\beta$ = 3.026e-04, SE = 8.245e-05, p = 0.000261, $\eta^2$ = 0.02), suggestive of ongoing learning or adaptation to crutch-assisted locomotion.

In the ML axis, in comparison to the rigid crutches, both spring-loaded ($\beta$ = 1.653e-03, SE = 5.859e-04, p = 0.0049, $\eta^2$ = 0.03) and tensegrity devices ($\beta$ = -1.421e-03, SE = 6.253e-04, p = <.0001) had significant lower impulse. Pairwise comparisons further indicated that the tensegrity condition resulted in lower ML impulse than the spring condition ($\beta$ = 0.00307, SE = 0.000620, p = 0.0234, $\eta^2$ = 0.03), see Figure 5d. ML impulse tended to decrease along consecutive block of trials ($\beta$ = -7.209e-04, SE = 3.088e-04, p = 0.0198, $\eta^2$ = 7.55e-03) and trials within block ($\beta$ = -1.985e-04, SE = 8.038e-05, p = 8.45e-03, $\eta^2$ = 0.03).

**Speed of Walking**

In straight walking, there was no effect of spring-loaded (p = 0.302) or tensegrity device type (p = 0.105), see Figure 6a, but there was an interaction between spring-loaded device and block ($\beta$ = -0.0387, SE = 0.0148, p = 0.009) and an effect of block ($\beta$ = 0.0538, SE = 0.0103, p = 0.006), indicating that speed decreased when walking with spring-loaded crutches later in the session. Pairwise comparisons showed marginally significant differences between device types, where speed was lower with spring-loaded types than with rigid (β = -0.020338, SE = 0.00854, p = 0.0461) and tensegrity types (β =-0.019714, SE = 0.00873, p = 0.0628).

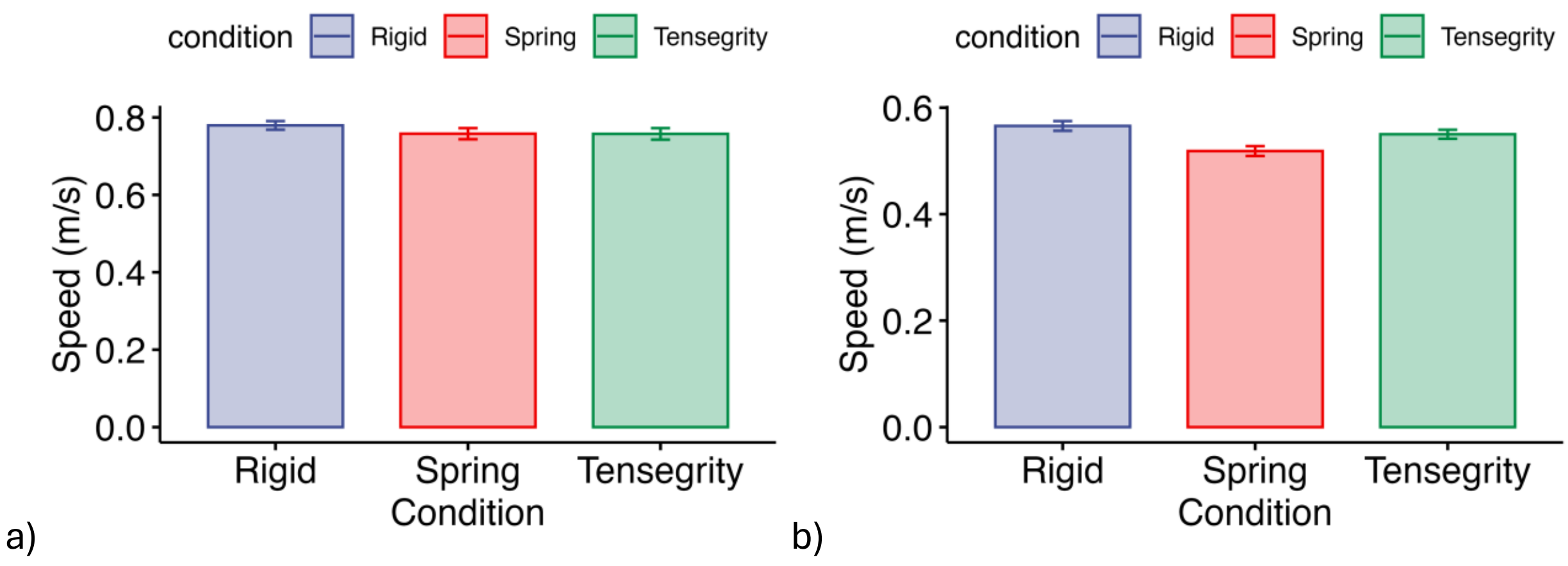

*Figure* 6. Walking speed (*M*±*SE*) per device type and path trajectory.

In trials involving turning, compared to rigid crutches, walking speed was significantly lower with spring-loaded crutches than rigid ones (β = -0.035886, SE = 0.006827, p = 2.75e-07), and there was no difference between rigid and tensegrity crutches (p = 0.494), see Figure 6b. Pairwise comparisons further indicated that the tensegrity device resulted in higher walking speeds than the spring-loaded type (β = -0.03117, SE = 0.00622, p<.0001). Walking speed also tended to increase along blocks of trials (β = 0.014004, SE = 0.003517, p = 8.52e-05).

**Perceived Effort, Pain, Comfort, Usability, and Stability**

***Effort***

Effort evaluated with the Borg scale was highest with the spring-loaded crutches, see Figure 7a. Effort was not significantly different between rigid and tensegrity crutches (β = 0.1921, SE = 0.1634, p = 0.242999). It was significantly higher with spring-loaded than tensegrity crutches (β = 0.5635, SE = 0.1634, p = 0.000887). Pairwise comparisons further indicated that there was no difference between rigid and spring-loaded condition (β = -0.371, SE = 0.162, p = 0.0617). No effects of or interactions with turning were observed (all p’s > .05).

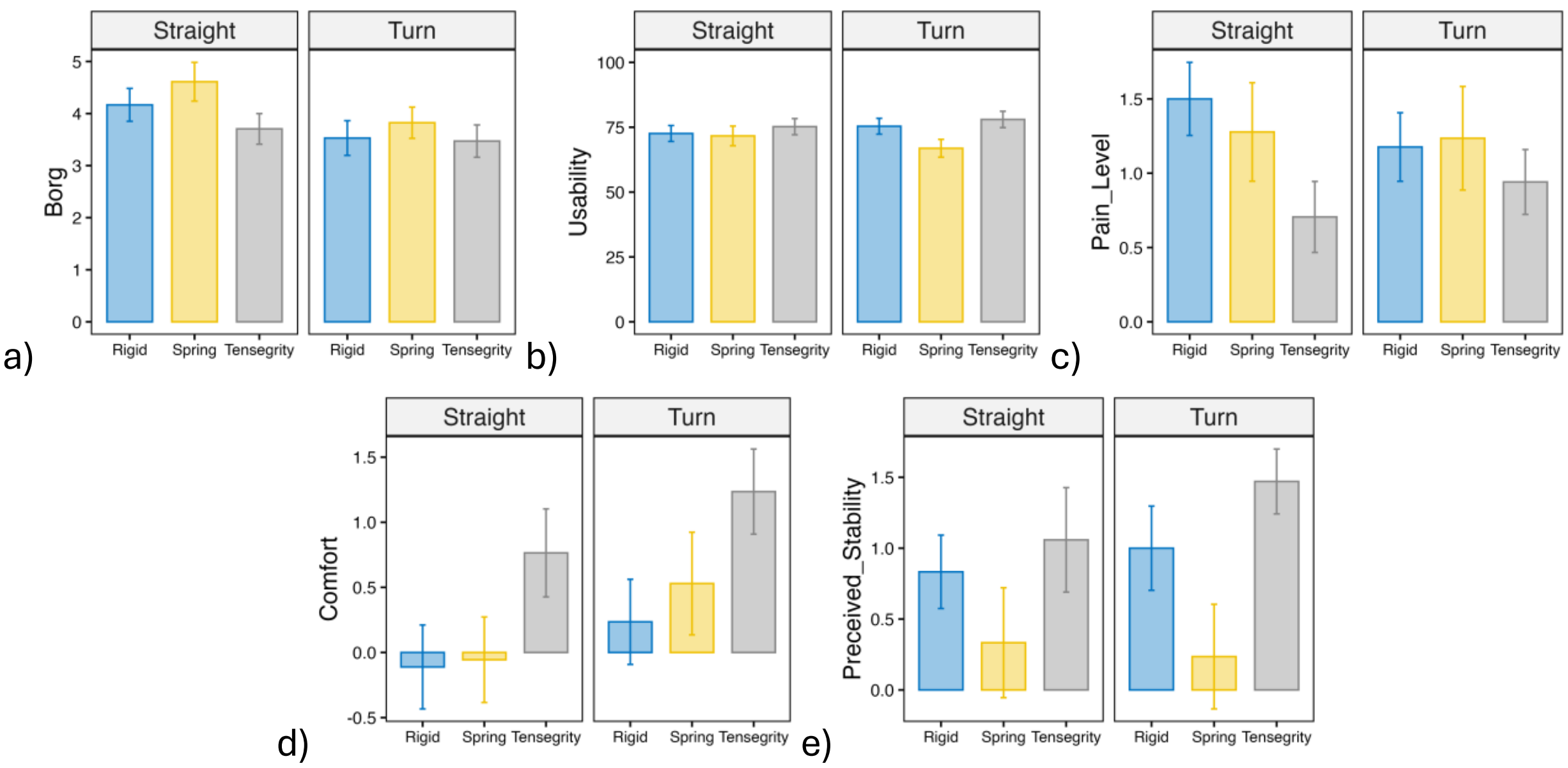

*Figure 7*. User evaluations of the crutches (*M*±*SE*). a) The Borg scale indicated higher effort with the spring-loaded crutches. b) The tensegrity type was significantly more usable than the spring-loaded, and there were no other effects. c) Pain level was highest with the rigid, followed by the spring-loaded, and it was lowest with the tensegrity module. d) The comfort level of the tensegrity design was higher than both the rigid and spring-loaded designs. e) The perceived stability of the spring-loaded crutches was significantly lower than that of the tensegrity type.

### ***Usability***

Usability ratings were significantly lower for spring-loaded than tensegrity crutches ($\beta$ = -6.613, SE = 2.617, p = 0.0133), while there was no significant difference between rigid and tensegrity types ($\beta$ = -2.470, SE = 2.617, p = 0.3478), see Figure 7b. Pairwise comparisons further indicated that there was no difference between rigid and spring-loaded types ($\beta$ = 4.14, SE = 2.59, p = 0.2526). No effects of or interactions with turning were observed (all p's > .05).

### ***Pain***

Compared to using tensegrity crutches, reported pain level was significantly higher after walking with the spring-loaded ($\beta$ = 0.4161, SE = 0.1419, p = 0.00432) and rigid crutches ($\beta$ = 0.5018, SE = 0.1419, p = 0.00066), see Figure 7c. Pairwise comparisons further indicated that there was no difference between the rigid and spring-loaded types ($\beta$ = 0.0857, SE = 0.140, p = 0.8148). No effects of or interactions with turning were observed (all p's > .05).

***Comfort***

Compared to tensegrity crutches, comfort level was significantly lower when using spring-loaded crutches (β = -0.7472, SE = 0.2442, p = 0.002975) and rigid crutches (β = -0.9186, SE = 0.2442, p = 0.000312), see Figure 7d. Pairwise comparisons further indicated that there was no difference between rigid condition and spring-loaded condition (β = -0.171, SE = 0.242, p = 0.7589). No effects of or interactions with turning were observed (all p’s > .05).

***Perceived Stability***

Reported stability was significantly lower when using spring-loaded crutches than tensegrity crutches (β = -0.9780, SE = 0.2882, p = 0.00105), and there was no significant difference between the rigid and tensegrity types (β = -0.3494, SE = 0.2882, p = 0.22876), see Figure 7e. Pairwise comparisons further indicated that there was no difference between rigid and spring-loaded types (β = 0.629, SE = 0.286, p = 0.0771). No effects of or interactions with turning were observed (all p’s > .05).

## Discussion

We developed and evaluated an innovative mobility aid with tensegrity-based compliance. The crutches with a pre-stressed self-tensile module combined several improvements in parameters measured during walking trials and in participants’ evaluations. Ground reaction force profiles developed more smoothly than when walking with rigid crutches. In the AP dimension, peak loading rate was even lower than with spring-loaded crutches. Expectedly, walking with spring-loaded crutches had higher impulse than with the other aid types in all dimensions. Participants’ evaluations favored the tensegrity crutches. Ratings of effort were lower for tensegrity crutches than spring-loaded. Comfort was highest with the tensegrity crutches, and pain was lowest. Usability of the tensegrity design was higher than the spring-loaded. Perceived stability was lower with the spring-loaded crutches, and rigid and tensegrity were not different.

To summarize, the tensegrity design achieved its main objective: it produced smoother forces in walking trials and it was more comfortable and less painful than the rigid design, although it was not necessarily perceived as more stable than the rigid. Importantly, these benefits were not at the expense of performance; walking speed was not different between rigid and tensegrity. The spring-loaded design also resulted in smoother forces than the rigid design, as expected, but it exhibited several disadvantages. It was perceived as less stable than the other two designs and resulted in more effort and impulse and lower walking speeds.

Comparing the tensegrity and spring-loaded crutches led to interesting insight. Spring-loaded crutches are designed to address some of the same issues raised here, namely, the high peak forces of rigid crutches that are likely to cause discomfort and secondary injury. Yet, this is achieved at the expense of higher effort, lower speed, and lower perceived stability. The mechanical testing that we conducted suggested a possible explanation. The spring-loaded crutches had a flat stiffness and long travel. Most participants sank by about 3 cm while loading in the stance phase. This likely interacted with balance control. Indeed, at debriefing several of the participants complained that the spring-loaded crutches required more effort to keep under control and felt bouncy. It is possible that this induced a more cautious gait which would explain the lower walking speed.

The tensegrity crutches had a steeper non-linear stiffness and short travel. This likely contributed to achieving smoothness without sacrificing comfort, stability, and speed. There are other beneficial properties of tensegrities that are relevant in this context, such as wider base of support, ground-following, and feedback. It is possible that damping is a factor too. Tensegrity structures do not inherently have high damping but they still have higher structural damping than pure elastic materials such as steel. In physically realized tensegrities, effects such as joint friction increase damping and make the structures better at dissipating unwanted vibrations (Oppenheim & Williams, 2001).

The present work contributes to a growing field of research with exciting theoretical and practical implications (Gomez-Jauregui et al., 2023). The tensegrity hypothesis is foundational for a new branch of biomechanics. It accommodates complexities of musculoskeletal anatomy that are ignored by traditional approaches which reduce human movement to models of stacked rigid segments and hinges (Marjaninejad & Valero-Cuevas, 2019; Oliveira & Skelton, 2009). Our empirical study shows that the theoretical concepts can be implemented in a flexible and strong device with immediate benefits for movement.

This was among the first demonstrations of a robust device that could be tested by humans performing realistic full-body tasks. This is an important achievement for a field which is rich in concepts and simulations but is only beginning to show that they can endure real use. In this sense, the tensegrity crutch is a relatively accessible platform for testing theoretical ideas, unlike exoskeletons and prosthetic devices which are costly and limit recruitment. This platform will continue to advance our knowledge about the biomechanical implications of tensegrities because it crosses the gap between biologically inspired robotics and rehabilitation technology. Last but not least, our modular design is advantageous because it makes it possible to experiment easily with different crutches and even other types of load-bearing mobility aids.

**Future Directions**

The development and evaluation of a novel mobility aid implementing a tensegrity module confirmed its anticipated beneficial properties. In the process, crutches proved to be an ideal platform for investigating how to transfer the exciting theoretical principles to assistive technology. When considering the promising features of tensegrities, it would be tempting to start working in a challenging domain of application such as exoskeletons or prosthetics. Yet, behind its apparent simplicity the crutch hides a great potential for experimental investigation. We found that prototyping was not limited by the minimal material costs. Furthermore, we found it relatively easy to recruit participants. Members of the general population could perform the task and provide meaningful feedback within one session of testing. Importantly, the same paradigm allows to continue testing exciting theoretical predictions and model how tensegrity mechanics interact with human biomechanics, motor control, and perception.

What are some other ways in which principles from tensegrity can improve assistive technology? In this study, participants reported how stable they felt, but this was hardly a precise test. The real benefits of multi-point independently suspended contact points and absorption of small perturbations should become apparent when balance is challenged, such as when walking on uneven and unpredictable terrain. Future testing in a suitable environment is necessary to evaluate this hypothesis. Another potential benefit that is worth exploring is energy recycling, an important theme in exoskeletons and prosthetics. In humans, tissues of the lower limbs have elastic properties that are exploited to store a portion of the gait cycle energy and promote efficient movement (Labonte & Holt, 2022). It is worth testing investigating how tensegrity modules, being passive nonlinear elastic elements, need to be tuned to bring such benefit.

One of the least exploited contributions of tensegrities to assistive technology is their potential to improve perception. While the mechanical stability and mobility of tensegrity structures are being investigated in robotics, there is very little empirical work to address how tensegrity mechanics could enhance haptic and kinesthetic perception. From a theoretical perspective, the relationship between a moving animal and its environment can be characterized as one of mutual constraint (Gibson, 1979). While an animal is walking, the animal uses the same organ, the muscle, to mechanically enact and sense the ground through instantaneous reaction forces and other mechanical interactions. This can be illustrated through perceiving a hand-held object. When holding an object and interacting with it to perceive its properties through haptic and kinesthetic perception, muscle co-contraction is beneficial in several relevant ways. It increases lower arm stiffness to improve mechanical coupling, improves near-instantaneous distal to proximal transfer of reaction forces, and increases muscle spindle firing to provide proprioceptive afferents (Carboni et al., 2021; Proske & Allen, 2019).

Under the tensegrity model, the whole body is an organ of action and perception because the pre-stressed network of muscular-connective-skeletal tissue is a long-distance, near zero-lag transducer of mechanical interactions (Turvey & Fonseca, 2014). Because a mobility aid such as a crutch takes over the functions of the lower extremities, it must adequately supplement not only stable load-bearing but also mechanical proprioceptive feedback. For a tool to be properly integrated as an extension of the body in the person-plus-object system (Vauclin et al., 2023), it probably needs to match natural compliance properties that the upper body can exploit for mechanical feedback. Our simulations provide preliminary confirmation that this was possible. The tensegrity module was subject to vertical loading from the top while stepping on an inclined surface. As Figure 2c shows, the top nodes where the rest of the crutch would be attached experienced specifically directed force vectors, and this did not interfere with the module's main role of supporting the vertical load without deforming catastrophically. This is further evidence that this biologically inspired design has the potential to satisfy the joint demands for mobility, stability, comfort, and sensory feedback in load-bearing assistive devices.

## Acknowledgements

DD received support from NIH P20GM109090 during preparation of this study. JG received support from NIH P20GM109090 and NIH 5P20GM152301-02.

## Appendix A: Statistical model tables

### 1) Vertical Peak Loading Rate

| | Model 0 | **Model 1** | Model 2 | Model 3 | Model 4 |
|---|---|---|---|---|---|
| Rigid (Baseline) | 3.9429*** | 5.7125*** | 5.5393*** | 5.7059*** | 5.8860*** |
| | (0.4193) | (0.4456) | (0.4904) | (0.5113) | (0.5437) |
| Spring | | -2.7347*** | -2.7230*** | -2.7290*** | -3.2485*** |
| | | (0.1817) | (0.1823) | (0.1823) | (0.3757) |
| Tensegrity | | -3.0931*** | -3.0505*** | -3.0514*** | -3.0158*** |
| | | (0.1854) | (0.1922) | (0.1922) | (0.3851) |
| Block | | | 0.0799 | 0.0754 | 0.0627 |
| | | | (0.0956) | (0.0957) | (0.0959) |
| Trial | | | | -0.0282 | -0.0552 |
| | | | | (0.0245) | (0.0401) |
| Spring*Trial | | | | | 0.0943 |
| | | | | | (0.0594) |
| Tensegrity*Trial | | | | | -0.0075 |
| | | | | | (0.0592) |
| AIC | 3508.2950 | 3232.3862 | 3236.5449 | 3242.8039 | 3251.2121 |
| BIC | 3522.1230 | 3255.4329 | 3264.2010 | 3275.0693 | 3292.6962 |
| Log Likelihood | -1751.1475 | -1611.1931 | -1612.2724 | -1614.4019 | -1616.6060 |
| Num. obs. | 742 | 742 | 742 | 742 | 742 |
| Num. groups: participant | 18 | 18 | 18 | 18 | 18 |
| Var: participant Rigid (Baseline) | 2.9985 | 3.2962 | 3.2761 | 3.2761 | 3.2919 |
| Var: Residual | 6.1144 | 4.1389 | 4.1412 | 4.1393 | 4.1305 |

***$p < 0.001$; **$p < 0.01$; *$p < 0.05$

**2) Anteroposterior Peak Loading Rate**

| | Model 0 | **Model 1** | Model 2 | Model 3 | Model 4 |
|---|---|---|---|---|---|
| Rigid (Baseline) | 1.3762*** | 1.8769*** | 1.8884*** | 1.9193*** | 1.9339*** |
| | (0.1396) | (0.1519) | (0.1708) | (0.1794) | (0.1923) |
| Spring | | -0.6916*** | -0.6924*** | -0.6935*** | -0.7558*** |
| | | (0.0683) | (0.0685) | (0.0686) | (0.1417) |
| Tensegrity | | -0.9667*** | -0.9695*** | -0.9696*** | -0.9399*** |
| | | (0.0697) | (0.0723) | (0.0723) | (0.1452) |
| Block | | | -0.0053 | -0.0061 | -0.0081 |
| | | | (0.0360) | (0.0360) | (0.0362) |
| Trial | | | | -0.0052 | -0.0071 |
| | | | | (0.0092) | (0.0151) |
| Spring*Trial | | | | | 0.0114 |
| | | | | | (0.0224) |
| Tensegrity*Trial | | | | | -0.0054 |
| | | | | | (0.0223) |
| AIC | 1955.9471 | 1783.0856 | 1789.8770 | 1799.0897 | 1814.3084 |
| BIC | 1969.7752 | 1806.1323 | 1817.5331 | 1831.3552 | 1855.7925 |
| Log Likelihood | -974.9736 | -886.5428 | -888.9385 | -892.5449 | -898.1542 |
| Num. obs. | 742 | 742 | 742 | 742 | 742 |
| Num. groups: participant | 18 | 18 | 18 | 18 | 18 |
| Var: participant Rigid (Baseline) | 0.3301 | 0.3758 | 0.3763 | 0.3766 | 0.3772 |
| Var: Residual | 0.7544 | 0.5851 | 0.5859 | 0.5864 | 0.5876 |

***$p < 0.001$; **$p < 0.01$; *$p < 0.05$

**3) Mediolateral Peak Loading Rate**

| | Model 0 | Model 1 | Model 2 | **Model 3** | Model 4 |
|---|---|---|---|---|---|
| Rigid (Baseline) | 0.9637*** | 1.0006*** | 0.8612*** | 0.9277*** | 0.9428*** |
| | (0.0824) | (0.0855) | (0.0967) | (0.1021) | (0.1101) |
| Spring | | -0.0871* | -0.0778 | -0.0802 | -0.1489 |
| | | (0.0412) | (0.0411) | (0.0410) | (0.0845) |
| Tensegrity | | -0.0312 | 0.0029 | 0.0025 | 0.0388 |
| | | (0.0420) | (0.0433) | (0.0432) | (0.0867) |
| Block | | | 0.0643** | 0.0625** | 0.0602** |
| | | | (0.0215) | (0.0215) | (0.0216) |
| Trial | | | | -0.0112* | -0.0131 |
| | | | | (0.0055) | (0.0090) |
| Spring*Trial | | | | | 0.0125 |
| | | | | | (0.0134) |
| Tensegrity*Trial | | | | | -0.0066 |
| | | | | | (0.0133) |
| AIC | 1023.1854 | 1031.9251 | 1030.8985 | 1037.3031 | 1053.1613 |
| BIC | 1037.0135 | 1054.9718 | 1058.5546 | 1069.5686 | 1094.6454 |
| Log Likelihood | -508.5927 | -510.9625 | -509.4492 | -511.6516 | -517.5806 |
| Num. obs. | 742 | 742 | 742 | 742 | 742 |
| Num. groups: participant | 18 | 18 | 18 | 18 | 18 |
| Var: participant Rigid (Baseline) | 0.1164 | 0.1175 | 0.1149 | 0.1153 | 0.1156 |
| Var: Residual | 0.2132 | 0.2124 | 0.2102 | 0.2093 | 0.2093 |

***$p < 0.001$; **$p < 0.01$; *$p < 0.05$

**4) Vertical Impulse**

| | Model 0 | Model 1 | **Model 2** | Model 3 | Model 4 |
|---|---|---|---|---|---|
| Rigid (Baseline) | $0.2909^{***}$ | $0.2820^{***}$ | $0.2948^{***}$ | $0.2985^{***}$ | $0.3007^{***}$ |
| | (0.0078) | (0.0079) | (0.0086) | (0.0090) | (0.0095) |
| Spring | | $0.0167^{***}$ | $0.0159^{***}$ | $0.0158^{***}$ | $0.0159^{**}$ |
| | | (0.0030) | (0.0030) | (0.0030) | (0.0062) |
| Tensegrity | | $0.0117^{***}$ | $0.0089^{**}$ | $0.0089^{**}$ | 0.0001 |
| | | (0.0031) | (0.0032) | (0.0032) | (0.0064) |
| Block | | | $-0.0059^{***}$ | $-0.0060^{***}$ | $-0.0059^{***}$ |
| | | | (0.0016) | (0.0016) | (0.0016) |
| Trial | | | | -0.0006 | -0.0011 |
| | | | | (0.0004) | (0.0007) |
| Spring*Trial | | | | | -0.0000 |
| | | | | | (0.0010) |
| Tensegrity*Trial | | | | | 0.0016 |
| | | | | | (0.0010) |
| AIC | -2809.9319 | -2818.7334 | -2820.0694 | -2806.6444 | -2781.5640 |
| BIC | -2796.1200 | -2795.7137 | -2792.4458 | -2774.4168 | -2740.1285 |
| Log Likelihood | 1407.9659 | 1414.3667 | 1416.0347 | 1410.3222 | 1399.7820 |
| Num. obs. | 738 | 738 | 738 | 738 | 738 |
| Num. groups: participant | 18 | 18 | 18 | 18 | 18 |
| Var: participant Rigid (Baseline) | 0.0011 | 0.0011 | 0.0011 | 0.0011 | 0.0011 |
| Var: Residual | 0.0012 | 0.0011 | 0.0011 | 0.0011 | 0.0011 |

$^{***}p < 0.001$; $^{**}p < 0.01$; $^{*}p < 0.05$

**5) Anteroposterior Propulsive Impulse**

| | Model 0 | Model 1 | Model 2 | **Model 3** | Model 4 |
|---|---|---|---|---|---|
| Rigid (Baseline) | 0.0174*** | 0.0166*** | 0.0152*** | 0.0134*** | 0.0135*** |
| | (0.0012) | (0.0013) | (0.0015) | (0.0015) | (0.0017) |
| Spring | | 0.0026*** | 0.0027*** | 0.0027*** | 0.0025 |
| | | (0.0006) | (0.0006) | (0.0006) | (0.0013) |
| Tensegrity | | -0.0001 | 0.0001 | 0.0002 | 0.0002 |
| | | (0.0006) | (0.0006) | (0.0006) | (0.0013) |
| Block | | | 0.0006 | 0.0007* | 0.0007* |
| | | | (0.0003) | (0.0003) | (0.0003) |
| Trial | | | | 0.0003*** | 0.0003* |
| | | | | (0.0001) | (0.0001) |
| Spring*Trial | | | | | 0.0000 |
| | | | | | (0.0002) |
| Tensegrity*Trial | | | | | -0.0000 |
| | | | | | (0.0002) |
| AIC | -5172.711 | -5167.268 | -5154.773 | -5149.163 | -5114.595 |
| BIC | -5158.899 | -5144.248 | -5127.149 | -5116.935 | -5073.159 |
| Log Likelihood | 2589.355 | 2588.634 | 2583.386 | 2581.581 | 2566.2977 |
| Num. obs. | 738 | 738 | 738 | 738 | 738 |
| Num. groups: participant | 18 | 18 | 18 | 18 | 18 |
| Var: participant Rigid (Baseline) | 0.0000 | 0.0000 | 0.0000 | 0.0000 | 0.0000 |
| Var: Residual | 0.0000 | 0.0000 | 0.0000 | 0.0000 | 0.0000 |

***$p < 0.001$; **$p < 0.01$; *$p < 0.05$

**6) Anteroposterior Braking Impulse**

| | Model 0 | **Model 1** | Model 2 | Model 3 | Model 4 |
|---|---|---|---|---|---|
| Rigid (Baseline) | 0.0091*** | 0.0094*** | 0.0099*** | 0.0102*** | 0.0104*** |
| | (0.0007) | (0.0008) | (0.0008) | (0.0009) | (0.0009) |
| Spring | | 0.0002 | 0.0002 | 0.0002 | -0.0001 |
| | | (0.0003) | (0.0003) | (0.0003) | (0.0006) |
| Tensegrity | | -0.0012*** | -0.0013*** | -0.0013*** | -0.0015* |
| | | (0.0003) | (0.0003) | (0.0003) | (0.0006) |
| Block | | | -0.0002 | -0.0002 | -0.0002 |
| | | | (0.0001) | (0.0001) | (0.0001) |
| Trial | | | | -0.0001 | -0.0001 |
| | | | | (0.0000) | (0.0001) |
| Spring*Trial | | | | | 0.0000 |
| | | | | | (0.0001) |
| Tensegrity*Trial | | | | | 0.0000 |
| | | | | | (0.0001) |
| AIC | -6297.956 | -6291.652 | -6276.089 | -6257.763 | -6220.346 |
| BIC | -6284.144 | -6268.632 | -6248.465 | -6225.535 | -6178.911 |
| Log Likelihood | 3151.978 | 3150.826 | 3144.045 | 3135.882 | 3119.173 |
| Num. obs. | 738 | 738 | 738 | 738 | 738 |
| Num. groups: participant | 18 | 18 | 18 | 18 | 18 |
| Var: participant Rigid (Baseline) | 0.0000 | 0.0000 | 0.0000 | 0.0000 | 0.0000 |
| Var: Residual | 0.0000 | 0.0000 | 0.0000 | 0.0000 | 0.0000 |

***p < 0.001; **p < 0.01; *p < 0.05

**7) Mediolateral Impulse**

| | Model 0 | Model 1 | Model 2 | **Model 3** | Model 4 |
|---|---|---|---|---|---|
| Rigid (Baseline) | $0.0298^{***}$ | $0.0295^{***}$ | $0.0310^{***}$ | $0.0322^{***}$ | $0.0321^{***}$ |
| | (0.0025) | (0.0025) | (0.0026) | (0.0027) | (0.0027) |
| Spring | | $0.0018^{**}$ | $0.0017^{**}$ | $0.0017^{**}$ | 0.0010 |
| | | (0.0006) | (0.0006) | (0.0006) | (0.0012) |
| Tensegrity | | -0.0011 | $-0.0014^{*}$ | $-0.0014^{*}$ | -0.0001 |
| | | (0.0006) | (0.0006) | (0.0006) | (0.0013) |
| Block | | | $-0.0007^{*}$ | $-0.0007^{*}$ | $-0.0008^{*}$ |
| | | | (0.0003) | (0.0003) | (0.0003) |
| Trial | | | | $-0.0002^{*}$ | -0.0002 |
| | | | | (0.0001) | (0.0001) |
| Spring*Trial | | | | | 0.0001 |
| | | | | | (0.0002) |
| Tensegrity*Trial | | | | | -0.0002 |
| | | | | | (0.0002) |
| AIC | -5194.793 | -5186.610 | -5175.195 | -5162.257 | -5130.389 |
| BIC | -5180.981 | -5163.591 | -5147.571 | -5130.029 | -5088.954 |
| Log Likelihood | 2600.397 | 2598.305 | 2593.597 | 2588.128 | 2574.195 |
| Num. obs. | 738 | 738 | 738 | 738 | 738 |
| Num. groups: participant | 18 | 18 | 18 | 18 | 18 |
| Var: participant Rigid (Baseline) | 0.0001 | 0.0001 | 0.0001 | 0.0001 | 0.0001 |
| Var: Residual | 0.0000 | 0.0000 | 0.0000 | 0.0000 | 0.0000 |

$^{***}p < 0.001$; $^{**}p < 0.01$; $^{*}p < 0.05$

**8) Speed. Straight condition.**

| | Model 0 | Model 1 | Model 2 | Model 3 | Model 4 | **Model 5** |
|---|---|---|---|---|---|---|
| Rigid (Baseline) | 0.7714*** | 0.7858*** | 0.7102*** | 0.6728*** | 0.6686*** | 0.6650*** |
| | (0.0387) | (0.0391) | (0.0403) | (0.0458) | (0.0463) | (0.0472) |
| Spring | | -0.0269** | -0.0247** | 0.0544 | 0.0545 | 0.0368 |
| | | (0.0090) | (0.0085) | (0.0321) | (0.0321) | (0.0357) |
| Tensegrity | | -0.0164 | -0.0033 | 0.0285 | 0.0285 | 0.0582 |
| | | (0.0091) | (0.0088) | (0.0317) | (0.0317) | (0.0359) |
| Block | | | 0.0357*** | 0.0534*** | 0.0535*** | 0.0538*** |
| | | | (0.0044) | (0.0104) | (0.0104) | (0.0103) |
| Spring*Block | | | | -0.0380* | -0.0380* | -0.0387** |
| | | | | (0.0148) | (0.0149) | (0.0148) |
| Tensegrity*Block | | | | -0.0148 | -0.0148 | -0.0160 |
| | | | | (0.0155) | (0.0155) | (0.0154) |
| Trial | | | | | 0.0007 | 0.0012 |
| | | | | | (0.0012) | (0.0020) |
| Spring*Trial | | | | | | 0.0035 |
| | | | | | | (0.0028) |
| Tensegrity*Trial | | | | | | -0.0048 |
| | | | | | | (0.0028) |
| AIC | -1008.917 | -998.505 | -1049.503 | -1038.774 | -1025.484 | -1009.900 |
| BIC | -996.082 | -977.112 | -1023.832 | -1004.546 | -986.977 | -962.837 |
| Log Likelihood | 507.459 | 504.253 | 530.751 | 527.387 | 521.742 | 515.950 |
| Num. obs. | 533 | 533 | 533 | 533 | 533 | 533 |
| Num. groups: pp | 18 | 18 | 18 | 18 | 18 | 18 |
| Var: pp Rigid (Baseline) | 0.0267 | 0.0268 | 0.0270 | 0.0284 | 0.0284 | 0.0284 |
| Var: Residual | 0.0074 | 0.0073 | 0.0065 | 0.0064 | 0.0064 | 0.0063 |

***$p < 0.001$; **$p < 0.01$; *$p < 0.05$

**9) Speed. Turning condition.**

| | Model 0 | Model 1 | **Model 2** | Model 3 | Model 4 | Model 5 |
|---|---|---|---|---|---|---|
| Rigid (Baseline) | 0.5482*** | 0.5683*** | 0.4919*** | 0.5083*** | 0.5037*** | 0.5005*** |
| | (0.0196) | (0.0199) | (0.0276) | (0.0486) | (0.0489) | (0.0490) |
| Spring | | -0.0472*** | -0.0359*** | -0.0323 | -0.0320 | -0.0373 |
| | | (0.0064) | (0.0068) | (0.0639) | (0.0639) | (0.0653) |
| Tensegrity | | -0.0163* | -0.0047 | -0.0673 | -0.0665 | -0.0550 |
| | | (0.0064) | (0.0069) | (0.0624) | (0.0624) | (0.0626) |
| Block | | | 0.0140*** | 0.0110 | 0.0111 | 0.0103 |
| | | | (0.0035) | (0.0081) | (0.0081) | (0.0081) |
| Spring*Block | | | | -0.0013 | -0.0013 | 0.0003 |
| | | | | (0.0123) | (0.0123) | (0.0123) |
| Tensegrity*Block | | | | 0.0128 | 0.0126 | 0.0141 |
| | | | | (0.0121) | (0.0121) | (0.0121) |
| Trial | | | | | 0.0011 | 0.0032 |
| | | | | | (0.0013) | (0.0021) |
| Spring*Trial | | | | | | -0.0009 |
| | | | | | | (0.0030) |
| Tensegrity*Trial | | | | | | -0.0052 |
| | | | | | | (0.0029) |
| AIC | -947.1470 | -978.9896 | -983.0400 | -966.4078 | -953.6119 | -933.3998 |
| BIC | -935.7771 | -960.0398 | -960.3003 | -936.0881 | -919.5022 | -891.7103 |
| Log Likelihood | 476.5735 | 494.4948 | 497.5200 | 491.2039 | 485.8059 | 477.6999 |
| Num. obs. | 327 | 327 | 327 | 327 | 327 | 327 |
| Num. groups: pp | 18 | 18 | 18 | 18 | 18 | 18 |
| Var: pp Rigid (Baseline) | 0.0068 | 0.0068 | 0.0068 | 0.0069 | 0.0070 | 0.0070 |
| Var: Residual | 0.0025 | 0.0021 | 0.0020 | 0.0020 | 0.0020 | 0.0020 |

$^{***}p < 0.001$; $^{**}p < 0.01$; $^{*}p < 0.05$

**10) Borg**

| | Model 0 | Model 1 | **Model 2** | Model 3 |
|---|---|---|---|---|
| Rigid (Baseline) | 3.9925*** | 3.9281*** | 4.1490*** | 4.1667*** |
| | (0.3080) | (0.3205) | (0.3200) | (0.3314) |
| Spring | | 0.3714* | 0.3714* | 0.4444* |
| | | (0.1714) | (0.1617) | (0.2261) |
| Tensegrity | | -0.1938 | -0.1921 | -0.3274 |
| | | (0.1732) | (0.1634) | (0.2308) |
| Turning | | | -0.4645*** | -0.5039* |
| | | | (0.1338) | (0.2308) |
| Spring*Turning | | | | -0.1503 |
| | | | | (0.3244) |
| Tensegrity*Turning | | | | 0.2686 |
| | | | | (0.3277) |
| AIC | 292.5690 | 289.5726 | 282.3063 | 285.7244 |
| BIC | 300.5022 | 302.7946 | 298.1726 | 306.8795 |
| Log Likelihood | -143.2845 | -139.7863 | -135.1531 | -134.8622 |
| Num. obs. | 104 | 104 | 104 | 104 |
| Num. groups: pp | 18 | 18 | 18 | 18 |
| Var: pp Rigid (Baseline) | 1.6033 | 1.5817 | 1.5331 | 1.5169 |
| Var: Residual | 0.5665 | 0.5142 | 0.4574 | 0.4601 |

***$p < 0.001$; **$p < 0.01$; *$p < 0.05$

**11) Comfort**

| | Model 0 | Model 1 | **Model 2** | Model 3 |
|---|---|---|---|---|
| Rigid (Baseline) | 0.3906 | 0.0366 | -0.1754 | -0.1111 |
| | (0.2554) | (0.2898) | (0.3022) | (0.3326) |
| Spring | | 0.1714 | 0.1714 | 0.0556 |
| | | (0.2471) | (0.2418) | (0.3408) |
| Tensegrity | | $0.9211^{***}$ | $0.9186^{***}$ | $0.8411^{*}$ |
| | | (0.2495) | (0.2442) | (0.3474) |
| Turning | | | $0.4439^{*}$ | 0.3117 |
| | | | (0.1997) | (0.3474) |
| Spring*Turning | | | | 0.2386 |
| | | | | (0.4890) |
| Tensegrity*Turning | | | | 0.1589 |
| | | | | (0.4936) |
| AIC | 352.8975 | 344.6351 | 343.1588 | 346.3794 |
| BIC | 360.8307 | 357.8571 | 359.0251 | 367.5345 |
| Log Likelihood | -173.4488 | -167.3176 | -165.5794 | -165.1897 |
| Num. obs. | 104 | 104 | 104 | 104 |
| Num. groups: pp | 18 | 18 | 18 | 18 |
| Var: pp Rigid (Baseline) | 0.9522 | 0.9572 | 0.9494 | 0.9463 |
| Var: Residual | 1.2297 | 1.0685 | 1.0234 | 1.0453 |

$^{***}p < 0.001$; $^{**}p < 0.01$; $^{*}p < 0.05$

**12) Pain**

| | Model 0 | **Model 1** | Model 2 | Model 3 |
|---|---|---|---|---|
| Rigid (Baseline) | 1.1707*** | 1.3594*** | 1.3706*** | 1.5000*** |
| | (0.2383) | (0.2502) | (0.2563) | (0.2673) |
| Spring | | -0.0857 | -0.0857 | -0.2222 |
| | | (0.1404) | (0.1412) | (0.1952) |
| Tensegrity | | -0.5018*** | -0.5018*** | -0.7637*** |
| | | (0.1419) | (0.1427) | (0.1993) |
| Turning | | | -0.0237 | -0.2931 |
| | | | (0.1168) | (0.1993) |
| Spring*Turning | | | | 0.2810 |
| | | | | (0.2801) |
| Tensegrity*Turning | | | | 0.5284 |
| | | | | (0.2830) |
| AIC | 252.4548 | 247.4299 | 251.8509 | 254.0430 |
| BIC | 260.3879 | 260.6519 | 267.7172 | 275.1981 |
| Log Likelihood | -123.2274 | -118.7150 | -119.9254 | -119.0215 |
| Num. obs. | 104 | 104 | 104 | 104 |
| Num. groups: pp | 18 | 18 | 18 | 18 |
| Var: pp Rigid (Baseline) | 0.9501 | 0.9474 | 0.9461 | 0.9432 |
| Var: Residual | 0.3932 | 0.3449 | 0.3490 | 0.3431 |

***p < 0.001; **p < 0.01; *p < 0.05

**13) Perceived Stability**

| | Model 0 | **Model 1** | Model 2 | Model 3 |
|---|---|---|---|---|
| Rigid (Baseline) | $0.8137^{***}$ | $0.9133^{***}$ | $0.8376^{**}$ | $0.8333^{**}$ |
| | (0.1857) | (0.2474) | (0.2726) | (0.3179) |
| Spring | | $-0.6286^{*}$ | $-0.6286^{*}$ | -0.5000 |
| | | (0.2858) | (0.2867) | (0.4027) |
| Tensegrity | | 0.3494 | 0.3479 | 0.2239 |
| | | (0.2882) | (0.2892) | (0.4096) |
| Turning | | | 0.1575 | 0.1651 |
| | | | (0.2360) | (0.4096) |
| Spring*Turning | | | | -0.2647 |
| | | | | (0.5779) |
| Tensegrity*Turning | | | | 0.2467 |
| | | | | (0.5827) |
| AIC | 364.3335 | 358.5829 | 361.1893 | 363.2107 |
| BIC | 372.2667 | 371.8049 | 377.0556 | 384.3658 |
| Log Likelihood | -179.1667 | -174.2915 | -174.5946 | -173.6054 |
| Num. obs. | 104 | 104 | 104 | 104 |
| Num. groups: pp | 18 | 18 | 18 | 18 |
| Var: pp Rigid (Baseline) | 0.3392 | 0.3638 | 0.3623 | 0.3592 |
| Var: Residual | 1.5912 | 1.4292 | 1.4385 | 1.4599 |

$^{***}p < 0.001$; $^{**}p < 0.01$; $^{*}p < 0.05$

**14) Usability**

| | Model 0 | **Model 1** | Model 2 | Model 3 |
|---|---|---|---|---|
| Rigid (Baseline) | 73.3452*** | 74.0249*** | 73.8554*** | 72.5889*** |
| | (1.9163) | (2.5376) | (2.7920) | (3.2261) |
| Spring | | -4.6069 | -4.6069 | -0.9267 |
| | | (2.8848) | (2.9013) | (4.0414) |
| Tensegrity | | 2.7469 | 2.7429 | 2.7827 |
| | | (2.9098) | (2.9265) | (4.1108) |
| Turning | | | 0.3507 | 2.9463 |
| | | | (2.3890) | (4.1108) |
| Spring*Turning | | | | -7.5769 |
| | | | | (5.7989) |
| Tensegrity*Turning | | | | -0.1663 |
| | | | | (5.8475) |
| AIC | 836.6205 | 826.5494 | 824.9537 | 816.2970 |
| BIC | 844.5537 | 839.7714 | 840.8201 | 837.4521 |
| Log Likelihood | -415.3103 | -408.2747 | -406.4769 | -400.1485 |
| Num. obs. | 104 | 104 | 104 | 104 |
| Num. groups: pp | 18 | 18 | 18 | 18 |
| Var: pp Rigid (Baseline) | 38.8921 | 40.6490 | 40.4505 | 40.3394 |
| Var: Residual | 153.4858 | 145.6353 | 147.3070 | 146.9995 |

***$p < 0.001$; **$p < 0.01$; *$p < 0.05$